\documentclass[letterpaper, 10 pt, conference]{ieeeconf}

\pdfoutput=1 

\let\labelindent\relax

\usepackage{graphics} 
\usepackage[pdftex]{graphicx}
\usepackage[font={small}]{caption}
\usepackage{subcaption}
\usepackage[rightcaption]{sidecap}

\usepackage{mathtools}
\usepackage{amsmath, amssymb, amscd}
\usepackage{bbm} 
\usepackage{amsfonts}
\usepackage{mathptmx}       
\DeclareMathAlphabet{\mathcal}{OMS}{lmsy}{m}{n}
\DeclareSymbolFont{largesymbols}{OMX}{cmex}{m}{n}
\usepackage{ textcomp } 

\usepackage{array} 
\usepackage{tabularx}
\usepackage{multicol}
\usepackage{multirow}

\usepackage{times} 
\usepackage{xspace}
\usepackage[english]{babel} 
\usepackage{fixltx2e} 
\usepackage{bm}
\usepackage{units}
\usepackage{setspace}

\usepackage{makeidx}
\usepackage{enumitem}
\usepackage[yyyymmdd,hhmmss]{datetime}
\usepackage[english]{babel}

\makeatletter
\let\NAT@parse\undefined
\makeatother
\usepackage[numbers]{natbib}

\usepackage[ruled,vlined,linesnumbered]{algorithm2e}

\usepackage{url}
\makeatletter
\g@addto@macro{\UrlBreaks}{\UrlOrds}
\makeatother
\usepackage{color}
\usepackage[usenames,dvipsnames,table,xcdraw]{xcolor}
\usepackage[T1]{fontenc}
\usepackage{csvsimple}

\newcommand{\todo}[1]{\textcolor{red}{[?#1?]}}
\newcommand{\tocite}[1]{\textcolor{red}{[cite]}}
\newcommand{\del}[1]{}
\newcommand{\etal}[1]{et al.}

\newcommand{\hirl}{HIRL\xspace}

\newcommand{\hirlfull}{Hierarchical Inverse Reinforcement Learning\xspace}

\newboolean{include-notes}
\setboolean{include-notes}{true}
\newcommand{\fp}[1]{\ifthenelse{\boolean{include-notes}}%
 {\textcolor{blue}{\textbf{FP: #1}}}{}}

\title{\LARGE \bf
HIRL: Hierarchical Inverse Reinforcement Learning \\
for Long-Horizon Tasks with Delayed Rewards
}

\author{%
Sanjay Krishnan, 
Animesh Garg, 
Richard Liaw,
Lauren Miller,
Florian T. Pokorny,
Ken Goldberg
\thanks{EECS \& IEOR, University of California, Berkeley CA USA; \texttt{\{sanjaykrishnan, animesh.garg,rliaw, ftpokorny, laurenm, goldberg\}@berkeley.edu}}%
}

\IEEEoverridecommandlockouts 

\begin{document}


\newtheorem{theorem}{Theorem}
\newtheorem{example}{Example}
\newtheorem{definition}{Definition}
\newtheorem{problem}{Problem}
\newtheorem{property}{Property}
\newtheorem{proposition}{Proposition}
\newtheorem{lemma}{Lemma}
\newtheorem{corollary}{Corollary}

\maketitle

\begin{abstract}
Reinforcement Learning (RL) struggles in problems with delayed rewards, and one approach is to segment the task into sub-tasks with incremental rewards.
We propose a framework called \hirlfull (\hirl), which is a model for learning sub-task structure from demonstrations.
\hirl decomposes the task into sub-tasks based on transitions that are consistent across demonstrations. These transitions are defined as changes in local linearity w.r.t to a kernel function~\cite{krishnan2015tsc}. 
Then, \hirl uses the inferred structure to learn reward functions local to the sub-tasks but also handle any global dependencies such as sequentiality.

We have evaluated \hirl on several standard RL benchmarks: Parallel Parking with noisy dynamics, Two-Link Pendulum, 2D Noisy Motion Planning, and a Pinball environment.
In the parallel parking task, we find that rewards constructed with \hirl converge to a policy with an 80\% success rate in 32\% fewer time-steps than those constructed with Maximum Entropy Inverse RL (MaxEnt IRL), and with partial state observation, the policies learned with IRL fail to achieve this accuracy while \hirl still converges.
We further find that that the rewards learned with \hirl are robust to environment noise where they can tolerate 1 stdev. of random perturbation in the poses in the environment obstacles while maintaining roughly the same convergence rate.
We find that \hirl rewards can converge up-to $6\times$ faster than rewards constructed with IRL.
\end{abstract}

\fontsize{10pt}{12.5pt}
\selectfont

\section{Introduction}
Reinforcement Learning (RL) is increasingly popular in robotics as it facilitates learning control policies through exploration~\cite{kober2013reinforcement, DBLP:conf/icra/LevineWA15,DBLP:conf/icml/SchulmanLAJM15,DBLP:conf/iros/HanLA15,DBLP:journals/corr/SchulmanHWA15,DBLP:journals/corr/ZhangLMFA15,finn2015learning,schulman2015high,DBLP:journals/corr/LevineFDA15,DBLP:journals/corr/FuLA15}.
However, it is well known that efficacy of RL algorithms is highly dependent on how reward functions are specified~\cite{DBLP:conf/icml/NgHR99}.
It is often the case that the reward is based on a quantity that is very difficult to directly optimize and is only observed well after the agent has made a decision.
For example, in surgical robotic suturing, the ultimate concern is the scar volume.
On the other hand, for the purposes of learning how to suture, it is more useful to consider the problem in shorter steps, e.g., ensure that the robot completes each of the stitches uniformly. 

\begin{figure}[t]
\centering
 \includegraphics[width=\columnwidth]{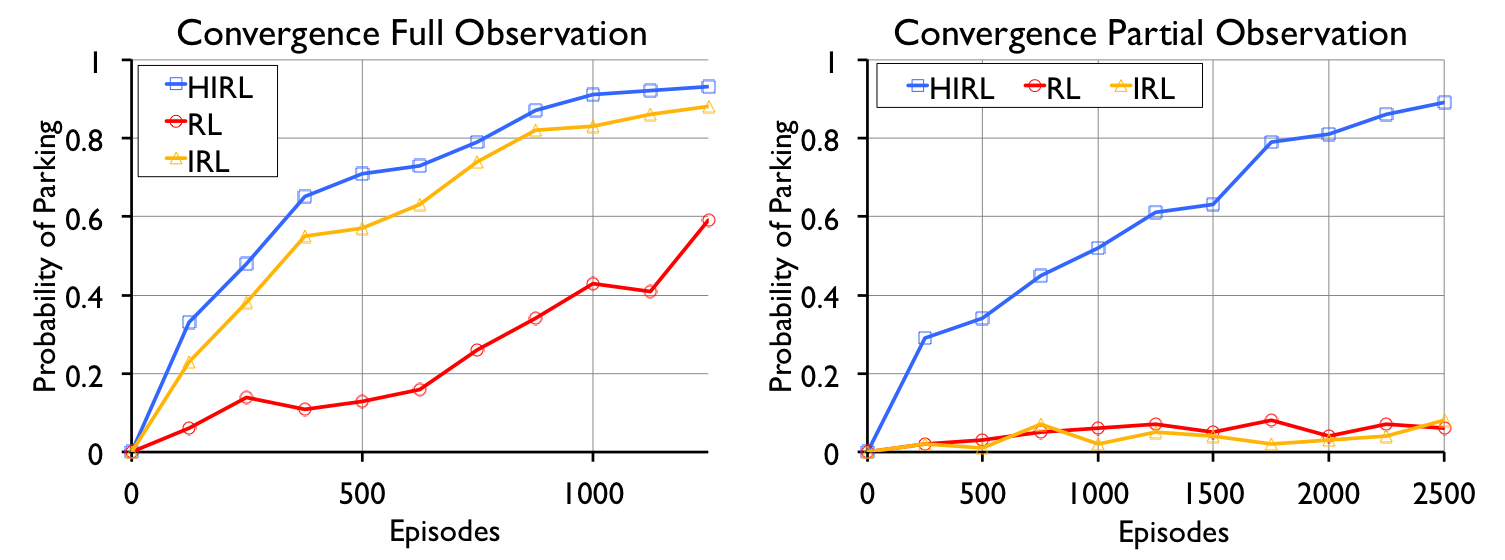}
 \caption{Convergence on a parallel parking task with noisy dynamics with full state observations (position, orientation, and velocity) and partial observation (only position and orientation); defining success as the probability that the car successfully parks.
 \hirl significantly reduces learning time in comparison to unsegmented RL and learns policies with a higher success rate than IRL on the same number of demonstrations especially under partial observation.\label{exp:rcsegmentation-res}}
 \vspace{-1em}
\end{figure}

It is often the case that complex tasks with delayed rewards can be segmented into a sequence of sub-tasks with shorter horizons.
Reward functions that capture this structure can lead to more efficient policy search.
This paper explores algorithms for learning such reward functions from a small number of supervisor demonstrations.
We build on our prior work of temporal segmentation~\cite{krishnan2015tsc,murali2016}, and propose the following model.
Let $D$ be a set of demonstrations $\{d_1,...,d_N\}$, each of which is a discrete time trajectory in some feature space $\mathbb{R}^p$ (e.g., a vector of joint angles).
Tasks are modeled as trajectories between \emph{transition states}, which are defined as states at which changes in local linearity that occur consistently across demonstrations.
This model is motivated by physical tasks where important events, like object contacts and forces/torques applied, are often correlated with changes in motion.
Identifying changes in local linearity is a substantially simpler problem than full system identification~\cite{willsky2009sharing,niekum2012learning, krishnan2015tsc,murali2016}, and thus, allows for efficient algorithms that can learn from relatively small datasets.

Once the sub-tasks are identified, simply placing rewards strategically at the end-points of segments can lead to inconsistencies since there is nothing enforcing the order of operations.
We show how we can avoid this problem by augmenting the state-space with additional variables that keep track of the previously reached transition states.
This can be modeled as using the segments as additional features in Inverse Reinforcement Learning (IRL, also called Inverse Optimal Control~\cite{boyd1994linear}).
With the additional features, the learned rewards will not only reflect the current state of the robot, but also the current active sub-task, and with the augmented states, any policy learning agent will be able to make use of these rewards.
We call the entire framework \hirlfull (\hirl), and it addresses two problems: (1) given a set of featurized demonstration trajectories, learn the locally linear sub-tasks, (2) use the learned sub-tasks to construct local rewards that respect the global sequential structure.

In this paper, we focus on a task hierarchy where tasks are composed from sequences of subtasks modeled by Linear-Gaussian systems. 

The general problem not only considers the dynamics within each subtask but transition dynamics between subtasks.
We formalize this inference problem in this paper, but defer the algorithmic discussion to future work.

Experiments on 7  RL benchmarks suggest that rewards constructed using \hirl can converge up-to 6x faster than those with  Maximum Entropy IRL~\cite{DBLP:conf/aaai/ZiebartMBD08}.
As a running example, consider a simulated parallel parking problem with noisy dynamics (Figure \ref{exp:rcsegmentation-res}).
There are two steps in parallel parking, which we call pulling-up and backing in.
This problem has delayed rewards because if the car does not perform the pulling-up step accurately, the car may miss the target when backing in.
It turns out that segmentation has an interesting side-effect in some problems under partial-observation.
Hypothetically, consider the case when the car cannot observe its own velocity.
For example, if the car's state-space is only defined in terms of $(x,y,\theta)$, the optimal policy has time-dependence the pulling-up and backing in steps different actions at similar points in the state-space.
If we partition the task sequentially into the two sub-tasks and ensured that they were independently executed, we could reduce the reward horizon and avoid the time-dependent policy.


\section{Preliminaries and Related Work}

\subsection{Background}\label{sec:back}
A finite-horizon Markov Decision Process (MDP) can be specified as $\mathcal{M} = \langle S,A,P(\cdot,\cdot), R(\cdot,\cdot),T \rangle$, where $S$ is the state-space, $A$ is the action space, $P: S \times A \mapsto Pr(S)$ is the transition function that maps states and actions to a probability density over subsequent states, $R: S \times A \mapsto \mathbb{R}$ is a reward function over the state and action space, and $T$ is the time-horizon.

Given some distribution over initial states $p_0$, an optimal policy $\pi^*$ is a policy that maximizes the expected reward:
\[
\pi^* = \arg\max_{\pi} \mathbf{E}_{d \sim \pi, p_0} [\sum_{t=0}^T R(s_t,a_t)],
\]
where $d \sim \pi, p_0$ denotes a distribution over the set of all \emph{trajectories} (sequences of state-action tuples) of length $T$ generated by the policy and initial conditions.

\subsection{Hierarchical RL}
Hierarchical RL (HRL) studies solving hierarchies of sub-problems posed as MDPs.
Our problem is a special case of learning hierarchies of MDPs as in HRL. 
The general problem not only considers the dynamics within each subtask but transition dynamics between subtasks.
We formalize this inference problem in this paper, but defer the algorithmic discussion to future work.
HRL considers a process, possibly stochastic, which transitions between a set of MDPs $\{\mathcal{M}_1,...,\mathcal{M}_k\}$. 

\subsubsection{Options}
Policies can constructed using regular actions and composite sequences of actions called ``options"~\cite{csimcsek2004using,menache2002q,mcgovern2001automatic}. These options can be defined a priori or they can be constructed through the process of exploration. 
The problem of discretizing the action-space is different from inferring local rewards for sub-tasks in \hirl.
Here we can make an analogy between policy learning and IRL in Learning From Demonstrations. Rewards are often argued to be more transferable, concise, and easier to interpret~\cite{DBLP:conf/icml/NgHR99}.
Similarly, decomposing MDPs in terms of sub-tasks with local rewards rather than in terms of composite actions may generalize better.

\subsubsection{Motion Primitives and Skill Learning}
Many of the ideas from HRL have been applied in robotics.
Motion primitives are segments that discretize the \emph{action}-space of a robot, and can facilitate faster convergence in LfD~\cite{ijspreet2002learning,pastor2009learning,manschitz2015learning}.
Furthermore, much of the initial work in motion primitives considered manually identified segments, but recently, Niekum et al. \cite{niekum2012learning} proposed learning the set of primitives from demonstrations using the Beta-Process Autoregressive Hidden Markov Model (BP-AR-HMM).
Calinon et al.~\cite{calinon2014skills} also build on a large corpus of literature of unsupervised skill segmentation including the task-parameterized movement model \cite{calinon2014task}, and GMMs for segmentation \cite{calinon2010learning}.

Recently, the robotics community has adopted some of the ideas from HRL in a field called skill learning \cite{konidaris2011robot}. Konidaris et al. studied largely the same problem proposed in this work, where demonstration trajectories are segmented into ``skills" using standard change point detection algorithms. These skills are used to build policies for complex RL tasks. Konidaris and Kuindersma et al.  studied many variants of this problem \cite{konidaris2009efficient}. 

\subsubsection{Sub-Tasks and State-Space Abstractions}
One of the earliest works in this field is by Kaelbling and Pack~\cite{kaelbling1993hierarchical}, where they proposed a technique to decompose a stochastic environment into Voronoi cells to improve learnability.
Dietrich et al. formalized idea of sub-task as an MDP~\cite{dietterich2000hierarchical}, and proposed an algorithm called MAXQ learning to address the information sharing problem.
McGovern and Barto studied this problem for discrete action and state-spaces where they identify states frequently visited by successful policies and use them to construct subgoals \cite{mcgovern2001automatic}. 
Kolter et al. also studied the problem called ``Hierarchical Apprenticeship Learning'' to learn bipedal locomotion~\cite{DBLP:conf/nips/KolterAN07}.
There is also some work in utlizing multi-task learning for RL~\cite{calandriello2014sparse}.
In \hirl, we explore how we can leverage demonstrations that are possibly spatially and temporally varying to infer such hierarchical structure.

\subsection{Inverse Reinforcement Learning}
In Inverse Reinforcement Learning (IRL) problems, we are given all of $\mathcal{M}$ except for the reward function, and a set of demonstrations $\mathcal{D}=\{d_1,...,d_N\}$ which are trajectories of the optimal policy $\pi^*$ with respect to some reward.
The objective is to infer $R(\cdot,\cdot)$ given the demonstrations~\cite{DBLP:conf/icml/NgHR99,DBLP:conf/aaai/ZiebartMBD08, coates2008learning}.
However, typically, it is impractical to observe enough data to learn the function $R(\cdot,\cdot)$ exactly.
Therefore, we often formulate the problem with parametrized reward functions:
\[
R_{\theta}(s,a) = \phi(f(s,a);\theta).
\]
where $f(s,a)$ is a feature vector in $\mathbb{R}^p$, $\theta$ is a parameter vector from some parameter space $\Theta \subseteq \mathbb{R}^q$, and $\phi$ describes the relationship between the two.
For example, we may restrict ourselves to the class of linear functions of the features:
\[
R_{\theta}(s,a) = f(s,a)^T\theta.
\]

Even with parametrization, the reward learning problem is often under-determined~\cite{DBLP:conf/icml/NgHR99}.
Differently shaped rewards can lead to different convergence rates when the rewards are used to learn policies in forward RL.
For IRL, the delayed reward problem has received some attention~\cite{DBLP:conf/ijcai/MacGlashanL15, DBLP:conf/aaai/JudahFTG14}, and in \hirl we consider sequential hierarchies of subtasks as well as reward learning.

\section{\hirl:\hirlfull}
In this section, we present the Hierarchical Inverse Reinforcement Learning model.

\subsection{\hirl Model}
We are given an MDP $\mathcal{M}$ with a known, but difficult to optimize, ``true'' reward function $R_{true}$ that is a binary indicator of success in some robotic task.
Let $D$ be a set of demonstrations $\{d_1,...,d_N\}$, we call these demonstrations \emph{expert} demonstrations if they are trajectories sampled from executions of the optimal policy $\pi^*$ with respect to $R_{true}$.
We assume that we are given featurization function $f:S\times A \mapsto \mathbb{R}^p$, and with this function, a demonstration is also a trajectory denoted by $x_t$ in $\mathbb{R}^p$.
The goal of \hirl is to construct a reward function $R_{seq}$ with shorter-term rewards whose optimal policy $\pi^\dagger$ approximates the performance of $\pi^*$.

We model $R_{true}$ in the following way.
Let $\rho \subseteq \mathbb{R}^p$ be subset of the state-space called a sub-goal.
A task is defined as an \emph{a priori} unknown sequence of sub-goals:
\[
G = [\rho_1,...,\rho_k]
\]
A task is successful, i.e., $R_{true}=1$, when all of the $\rho_i \in G$ are reached in sequence.
\hirl proposes an algorithm to learn $G$ in the case when the regions $\rho_i$ correspond to changes in local linearity, and following from $G$, $R_{seq}$ can be represented as a sequence of local rewards $[R^{(1)}_{seq},...,R^{(k)}_{seq}]$. 
The local reward sequence will serve to guide the agent to each of the $\rho_i$ more efficiently than the sparse true reward $R_{true}$.

\subsection{Locally Linear Sub-goals}
Consider the agent's trajectory in $\mathbb{R}^p$ as a dynamical system,
\[
x_{t+1} = \mathcal{T}(x_{t}) + w_{t},
\]
with i.i.d unit variance Gaussian process noise.
We model sub-tasks as locally-linear, that is, that the system $\mathcal{T}$ can be decomposed into a set of state-dependent linear systems:
\[
x_{t+1} = A_{i}\mathbf{x}_t + w_{t} \text{ : } A_i \in \{A_1,...,A_m\}.
\]
\emph{Transitions} are defined as times where $A_{t} \ne A_{t+1}$.
Thus, each transition will have an associated feature value $\mathbf{x}_{t}$ called a transition state.
The key insight from our prior work~\cite{krishnan2015tsc} is that the transition states have a meaningful spatial structure.
For example, we model these $\mathbf{x}_{t}$ as generated from a Gaussian Mixture Model (GMM) over the feature space $\mathbb{R}^p$.
We assume that the mixture components are separable~\cite{dasgupta2000two}.

We interpret this mixture model as defining sub-goals for the task.
If there $k$ mixture components for the distribution $\{m_1,...,m_k\}$, the quantile of each component distribution will define sequence of regions $[\rho_1,...,\rho_k]$ over the feature space (i.e., its sublevel set bounded by $z_\alpha$ and ordered by time), and can equivalently be thought of as $\rho_i = (\mu_i,\Sigma_i)$.
We interpret the learned $G = [\rho_1,...,\rho_k]$ as the sub-goals reached by the expert demonstrations.
Reaching transition states are associated with attaining rewards in the task, and when all of the transition states in $G$ are reached in sequence the agent is successful. 
Our Appendix contains details about this approach and intuition on where such a model may arise (Section \ref{sec:appendix1}).

\subsection{State-Space Augmentation}
Since, we are decomposing the task into a sequence of sub-goals, it leads to a problems that can no longer be modeled as an MDP.
Attaining a reward at goal $\rho_i$ depends on knowing that the reward at goal $\rho_{i-1}$ was attained.
This sequential dependence problem can arise even if the original problem is an MDP.
To model this dependence on the past, we have to construct an MDP whose state-space also includes history.
At first glance, this may seem impractical, but we can show that leveraging $G$ to compress this prior history leads to tractable learning problem.

Given a finite-horizon MDP $\mathcal{M}$ as defined in Section \ref{sec:back}, we can define an MDP $\mathcal{M}_H$ as follows.
Let $\mathcal{H}$ denote set of all dynamically feasible sequences of length $\le T$ comprised of the elements of $S$.
Therefore, for an agent at any time $t$, there is a sequence of previously visited states $H_t \in \mathcal{H}$.
The MDP $\mathcal{M}_H$ is defined as:
\[
\mathcal{M}_H = \langle S \times \mathcal{H},A,P'(\cdot,\cdot), R_\theta(\cdot,\cdot),T \rangle.
\]
For this MDP, $P'$ not only defines the transitions from the current state $s \mapsto s'$, but also increments the history sequence $H_{t+1} = H_{t} \sqcup s$.
Accordingly, the parametrized reward function $R_\theta$ is defined over $S$, $A$, and $H_{t+1}$.

By modeling assumption, we know that a sufficient statistic for task success is knowing that all of the transition states $G$ were reached.
We can use this fact to concisely encode the history of the agent $H_t \in \mathcal{H}$ in terms of transition states previously which is a $k$ dimensional vector $\{0,1\}^k$.
Then, additional complexity of representing the reward with history over $S \times  \{0,1\}^k$ is only $\mathbf{O}(k)$ instead of exponential in the time horizon.

\subsection{Reward Learning and Policy Evaluation}
With this model and the state-space augmentation, we can apply standard techniques for inverse and ``forward" RL.

\vspace{0.5em} \noindent \textbf{Inverse Reinforcement Learning: } We can apply standard techniques for IRL over the augmented state-space $S \times  \{0,1\}^k$. Suppose, we are considering linear functions of features of state-action tuples $R_{\theta}(s,a) = f(s,a)^T\theta$, we can apply the same reasoning to state-action-segment tuples $R_{\theta}(s,a) = \binom{f(s,a)^T}{v}\theta$, where $v \in \{0,1\}^k$ and indicates the sub-goal progress. Then, conditioned on each possible $v$ (i.e., the current task progress), we can find the local reward sequence $[R^{(1)}_{seq},...,R^{(k)}_{seq}]$. 
In principle, we can apply any IRL technique, and in this work, we apply the widely used Maximum Entropy IRL~\cite{DBLP:conf/aaai/ZiebartMBD08}.

\vspace{0.5em} \noindent \textbf{Rewards to Policies: }
In principle, we can apply many different policy learning techniques to learn a policy given the reward function over the augmented state-space $S \times  \{0,1\}^k$. In this paper, we use Q-learning to address the policy learning problem, and evaluate the claim of whether \hirl rewards lead to faster convergence. However, one could also apply these learned rewards in a framework like Guided Policy Search~\cite{DBLP:journals/corr/LevineFDA15}, or even in an optimal control framework like iLQR~\cite{li2004iterative}.


\section{Transition State Identification Algorithm}\label{sec:algo}
Now, we describe the transition state identification algorithm, which is derived from our prior work~\cite{krishnan2015tsc,murali2016}.

\subsection*{Step 1. Featurization}
The first step is to apply $f$ to every state-action pair in the demonstrations.
Every $d_i$ in $D$ is a sequence of $T$ state-action pairs:
\[
[(s_1,a_1),...,(s_T,a_T)].
\]
For each state-action pair, we apply the featurization $[ f(s_1,a_1),...,f(s_T,a_T)]$.
This gives us a $T$-step trajectory in the feature space $\mathbb{R}^p$ which we denote as $\mathbf{x}_t$.

\subsection*{Step 2. Finding Transitions}
Each demonstration, $\mathbf{x}_t$, is a trajectory in $\mathbb{R}^p$.
The key idea is to be able to detect switches in a noisy system.
There are a number of different techniques in the change-point detection literature~\cite{harchaoui2009kernel} that use kernels in time-series, however these do not often consider systems with dynamics.
One technique that has empirically found a lot of success in robotics has been to linearize non-linear dynamics with a GMM model~\cite{moldovan2013dirichlet,calinon2014task, khansari2011learning}.
It can be formally shown that a GMM model is equivalent to Bayesian Weighted Local Linear Regression to approximate a system $x_{t+1} = \mathcal{T}(x_{t}) + \epsilon$~\cite{moldovan2013dirichlet}.
To make this paper self-contained, we provide justification for this method with proofs in our Appendix~(Section \ref{sec:appendix2}).

In terms of intuition, consider the following system with a non-linear $\mathcal{T}$:
\[
x_{t+1} = \mathcal{T}(x_{t}) + w_{t}
\]
We could model this in a probabilistic way, where there is some joint probability density $p$ over both $x_{t}$ and $x_{t+1}$. 
Since the function is non-linear, the joint distribution $p$ can be very complex.
We choose to model $p$ as a GMM:
\[
p(x_{t},x_{t+1}) \sim GMM(k)
\]
This has the interpretation of defining locally linear dynamics, since conditioned on one of the mixture components, the conditional expectation $\mathbf{E}[x_{t+1} \mid  x_{t}]$ is linear.

In typical GMM formulations, we have to select the number of mixture components $m$ before hand.
However, we can apply results in Bayesian non-parametric statistics and jointly solve for the component locations and the number of components with an algorithm called DP-GMM~\cite{kulis2011revisiting} with a soft prior over the number of clusters\footnote{We use the default settings in \url{https://pypi.python.org/pypi/dpcluster}}.

The identification procedure is summarized in Algorithm \ref{alg:tsh1}.
We construct a data matrix $\Gamma$ which is the set of $\gamma(t)$ over all demonstrations and times, where $\gamma(t)$ defines a window of $l$ time-steps.
We apply DP-GMM to $\Gamma$ and map each $\gamma(t)$ to a most likely mixture component, and thus for each demonstration $i$ we get $\gamma_{index}^{(i)}(t)$ which gives us the cluster index at time $t$.
Then, we identify all of the times $t$ such that $\gamma_{index}^{(i)}(t) \ne \gamma_{index}^{(i)}(t+1)$, and the final result is a set $\Theta$ of tuples of demonstration id $i$ and time $t$.
In prior work, we found that this procedure is robust to noise and can scale well to higher dimensions~\cite{krishnan2015tsc,murali2016}.

\begin{algorithm}[t]
\small
\DontPrintSemicolon
\caption{Transition Identification \label{alg:tsh1}}
\KwData{Set of demonstrations:$\mathcal{D}$}

$\Gamma \leftarrow \emptyset$

\ForEach{$d_i \in \mathcal{D}$}{
   \ForEach{$t \in 0,1,...,T_i$}{
        $\gamma(t) = [\mathbf{x}_{t-k+1},...,\mathbf{x}_{t}]^T$
        
        $\Gamma = \Gamma \cup \gamma(t)$
   }
}
$\{\gamma_{index}^{(1)}(t),...,\gamma_{index}^{(N)}(t)\} \leftarrow$ \texttt{DP-GMM($\Gamma$)}
\vspace{0.5em}

$\Theta \leftarrow \emptyset$

\ForEach{$d_i \in \mathcal{D}$}{
   \ForEach{$t \in 0,1,...,T_i$}{
        \If{$\gamma_{index}^{(i)}(t) \ne \gamma_{index}^{(i)}(t+1)$}{
        $\Theta = \theta \cup (i,t)$
        }
   }
}

\KwResult{The set of transitions $\Theta$}
\end{algorithm}

\subsection*{Step 3. Correspondence}
Across all demonstrations, the set of transition states induces a density over the feature-space and time.
Intuitively, when an agent enters certain regions of the state-space at certain times, there is a propensity to switch.
We are interested in aggregating nearby (spatially and temporally) transition states together.
We model this density as a Gaussian Mixture Model with $k$ mixture components $\{m_1,...,m_k\}$.
As before, we learn this with DP-GMM in the feature space to find these clusters.

Each of the mixture components is a multivariate Gaussian distribution with some mean and covariance.
Individually, each of the mixture components is a Gaussian distribution, and defines a region of the feature space and a time interval.
Thus, the result is exactly the set of target goal regions:
\[
G = [\rho_1, \rho_2,...,\rho_m].
\]
The overall procedure is summarized in Algorithm \ref{alg:tsh2}.

\begin{algorithm}[t]
\small
\DontPrintSemicolon
\caption{Transition State Clustering \label{alg:tsh2}}
\KwData{Set of transitions:$\Theta$}

$Y \leftarrow \emptyset$

\ForEach{$(i,t) \in \Theta$}{
   $Y \leftarrow Y \cup x^{(i)}_{t}$
}
$G = [\rho_1, \rho_2,...,\rho_m]\leftarrow$ \texttt{DP-GMM(Y)}

\KwResult{The set of transition state clusters $G$}
\end{algorithm}

\subsection{Embedding For Segmentation}\label{sec:kern}
The interesting part about the proposed model is that we do not need to learn the underlying dynamical parameters to identify $G$; we only have to detect that the locations at which the system's dynamics have switched.
This is crucial when we have a small number of demonstrations because we avoid the data requirements of full system identification and can solve a substantially simpler transition detection problem~\cite{willsky2009sharing,niekum2012learning, krishnan2015tsc,murali2016}.
However, it all seems to rest on a seemingly strong assumption about local-linearity.
We can relax this assumption with a kernel embedding of the trajectories.

Let $\Omega = \{\omega_i\}$ be an indexed set of all $\omega_t$ over all demonstrations.
Let $\mathbf{\kappa}(\omega_0,\omega_1)$ define a kernel function over the set $\Omega$.
For example, if $\mathbf{\kappa}$ is the radial basis function (RBF), then:
$ \mathbf{\kappa}(\omega_0,\omega_1) = e^{\frac{-\|\omega_0-\omega_1\|_2^2}{2\sigma}}$.
$\mathbf{\kappa}$ naturally defines a matrix $M$ where:
\[
M_{ij} = \mathbf{\kappa}(\omega_i,\omega_j). 
\]
The top $p'$ eigenvalues define a new embedded feature vector for each $\omega$ in $\mathbb{R}^{p'}$.
In this embedded space, we can apply our transition identification procedure.
This procedure allows us to model non-linearities and states with different scaling properties.

\section{Reward Learning Algorithm}\label{sec:reward}
The next problem is to use the learned goals $G$ to construct rewards for the task.

\subsection{Augmentation}
The partitioning of a task requires understanding how the sub-tasks are sequentially coupled.
Therefore, during learning, the agent needs to be aware of its overall global progress.
We will show that we have to also include a vector of additional states $v$ as a state-space $\binom{s}{v}$ to account for these constraints.

The key idea is to use $G$ to concisely encode the history of process until a time $t$ in terms of previously completed sub-tasks. 
The result will be stored in a vector $v$, which the agent can use in forward RL.
Algorithm \ref{alg:tsh3} summarizes this process.
$v = e(H_t)$ is the vector that we use to augment the state-space of the RL problem.

\begin{algorithm}[t]
\small
\DontPrintSemicolon
\caption{Transition State Encoding \label{alg:tsh3}}
\KwData{Set of transition state clusters:$G$}
\KwData{Sequence of previously visited states: $H_t$}

$e \leftarrow [0,...,0]$

\ForEach{$(x,t) \in H_t$}{
   \If{$(x,t) \in$  \texttt{conf}(G)}{
      $i \leftarrow$ \texttt{find}(G,(x,t))
      
      $e[i] \leftarrow 1 \text{ if } e(i-1) = 1 \text{ or } i=0$
   }
}

\KwResult{Vector indicating which transition states where previously visited $v$}
\end{algorithm}

\subsection{Maximum Entropy Inverse Reinforcement Learning}
We use a technique called Maximum Entropy Inverse Reinforcement Learning (MaxEnt-IRL). 
MaxEnt-IRL uses the principle of maximum entropy, i.e., given some testable property select the maximum entropy distribution that encodes that property, to formalize the IRL problem.
For the IRL problem, this results in the following model.

\begin{figure*}[t]
\centering
 \includegraphics[width=0.8\textwidth]{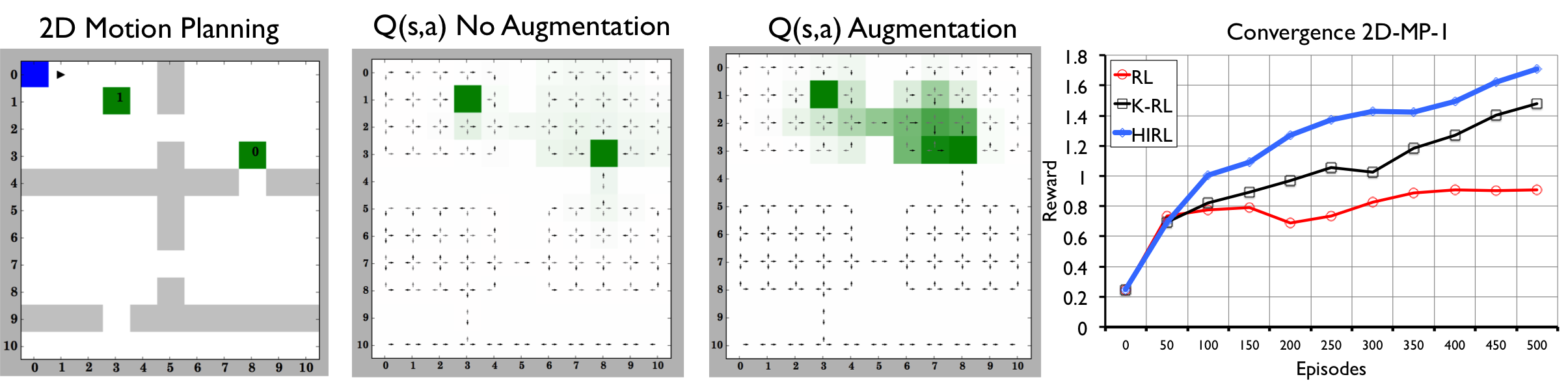}
 \caption{We constructed a scenario where an agent has to collect reward 0 and 1 in sequence. Since the agent has to cross over the same state in multiple stages of the task, RL can fail to converge in sequential tasks unless the state-space represents previous progress. 
 One approach is to segmented the problem into independent subtasks, which neglects any shared structure between the tasks (states far away from both goals are not valuable).
 \hirl augments the state-space with the appropriate variables and uses IRL to learn rewards that capture this structure. We plot the domain, the Q functions of the RL agent at convergence, and the learning curve (Section \ref{exp:2dmp}).  \label{exp:gweasy1}}
\end{figure*}

The observed data are modeled as trajectories $d_i$, and each possible trajectory is generated with probability:
\[
P(d_i | R) \propto \exp \{ \sum_{t=0}^T R(s_t,a_t) \}
\]
The distribution takes this form since given a fixed mean, the exponential distribution has the maximum entropy.
 MaxEnt-IRL uses the following linear parametrized representation:
\[
R(s,a) = f(s,a)^T \theta
\]
where $f(s,a)$ is the same feature vector representation used as before. 
The resulting form is:
\[
P(d_i | R) \propto \exp \{ \sum_{t=0}^T f(s_i,a_i)^T \theta \}
\]
and MaxEnt-IRL proposes an algorithm to infer the $\theta$ that maximizes the posterior likelihood.
To utilize segmentation, we propose the following variant:
\[
P(d_i | R) \propto \exp \{ \sum_{t=0}^T f(s_i,a_i)^T \theta_f  + v^T \theta_s\}
\]
which also incorporates the segments into the feature representation.
Thus, it jointly learns a parameter $\theta = \binom{\theta_f}{\theta_s}$ over both the segments and the features.


\subsection{Benefits of Knowing Transition States in IRL}
Now, we highlight some of the benefits of knowing segments when designing rewards.
First, there are tasks that are inherently sequential such as assembly.
For such tasks, there is a natural notion of sub-goals (i.e., the assembly of all of the components), and it is clear that a segmented reward model is required to learn optimal policies, since the agent needs to know previously finished sub-tasks.
A similar argument is clear for problems with partial observation, where some important states are not seen.
Knowing previously traversed states can disambiguate optimal actions.
Segmentation is one way to concisely encode the process history to allow for history dependent policies.
Surprisingly, we find that fitting such a reward model can lead to rewards that converge faster in forward RL (Section \ref{sec:exp})--even over techniques such as classical IRL, and we provide some intuition on why this can be the case.

\subsubsection{Simpler Local Policies} The additional segment features $v$ can also simplify policies leading to faster convergence. Consider the case, where the optimal policy is piecewise constant for each task segment. While this policy is easy to describe in terms of the features $v$, it may can be difficult to model in terms of the state-space $S$.

\subsubsection{Predictable Recovery} For more complex tasks, the agent will likely encounter states not seen in the set of demonstrations $D$, which will not be reflected in the reward function. In this case, the agent will explore until it arrives at known states and continue.
The additional features $v$ that track the segment progress encourage the agent to recover to the next sub-goal.
On the other hand, without the additional features IRL can miss sub-goals, leading to more unseen states in the future.
We find that when the number of demonstrations is relatively small (e.g, $5$) rewards constructed with \hirl (IRL with segment features) converges faster than IRL alone.



\section{Experiments}\label{sec:exp}
We evaluate \hirl in a series of standard RL benchmarks. Each experimental scenario follows a similar pattern: (1) we generate demonstration trajectories of the task, (2) apply \hirl to learn a reward function, (3) and compare how quickly a Q-learning RL agent converges to a solution with rewards learned with \hirl compared to other techniques.
In principle, the IRL problem (which learns rewards) is orthogonal to the problem of policy search (turning rewards into policies).
One could also use the same reward functions in optimal control frameworks such as iLQR, and we hope to explore this in further detail in future work. 

\subsection{Metrics}
For efficacy, we measure the max expected reward achieved by the agent (i.e., maximum over the entire learning epoch where we evaluate the policy after each episode and take the expected reward). For convergence rate, we measure the Area Under Curve of the learning curve (i.e., cumulative expected reward accrued over the entire learning epoch).
We compare against against IRL and RL directly on a default ``natural" reward function; which was defined as binary success or failure in discrete tasks and the squared distance to rewards in the continuous tasks.
We also include a comparison with direct policy learning from the demonstrations with a multi-class SVM\footnote{\url{http://scikit-learn.org/stable/modules/multiclass.html}} where applicable.

\subsection{2D Discrete Motion Planning}\label{exp:2dmp}
We modified a variant of one of the canonical RL domains, GridWorld, to illustrate how \hirl addresses problems of sequentiality (Figure \ref{exp:gweasy1}). 
This experiment intuitively illustrates the challenge of sequential rewards in RL.
We constructed a grid world with two goal states denoted by ``0'' and ``1'' separated by a narrow passage.
The agent can only receive the reward at ``1'' if it has previously reached ``0".
In the natural $(x,y)$ state-space, the agent does not learn a correct stationary policy since at some states the optimal action depends on knowing whether ``0'' has been reached.
We visualize the Q function of applying RL without state-space augmentation and we find that it predictably struggles in the narrow passage.

In the next figure, we show that augmenting the state-space with sub-task progress learned by \hirl from 5 demonstrations results in a Q function that reflects the sequential nature of the task.
For this task, we find that with \hirl, we can successfully learn a policy that has the desired sequential behavior.
Figure \ref{exp:gweasy1} also shows that \hirl converges faster than solving independent RL problems (K-RL).

\begin{figure}[t]
\centering
 \includegraphics[width=\columnwidth]{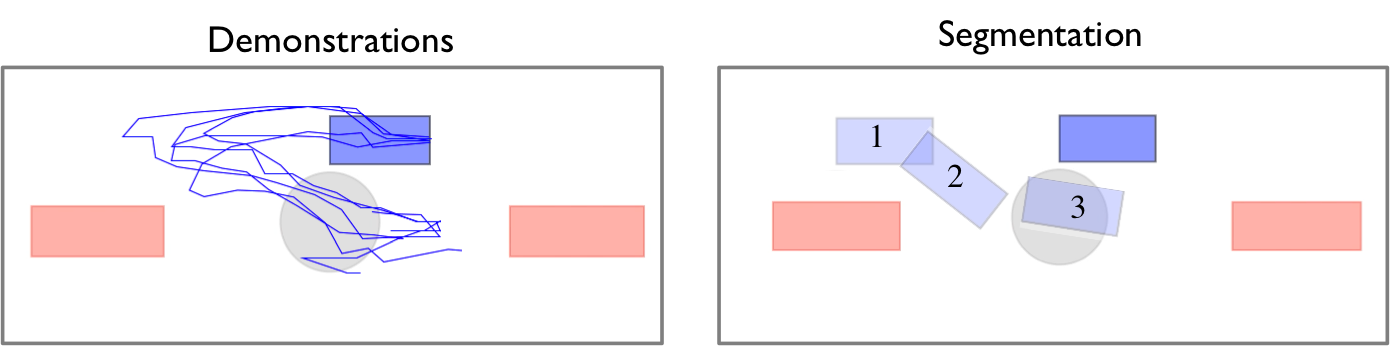}
 \caption{This plot illustrates the 5 demonstration trajectories for the parallel parking task, and the sub-goals learned by \hirl. Section \ref{exp:pp} describes the details of the experimental setup and the task. \label{exp:rcsegmentation}}
\end{figure}

\begin{figure}[t]
\centering
 \includegraphics[width=\columnwidth]{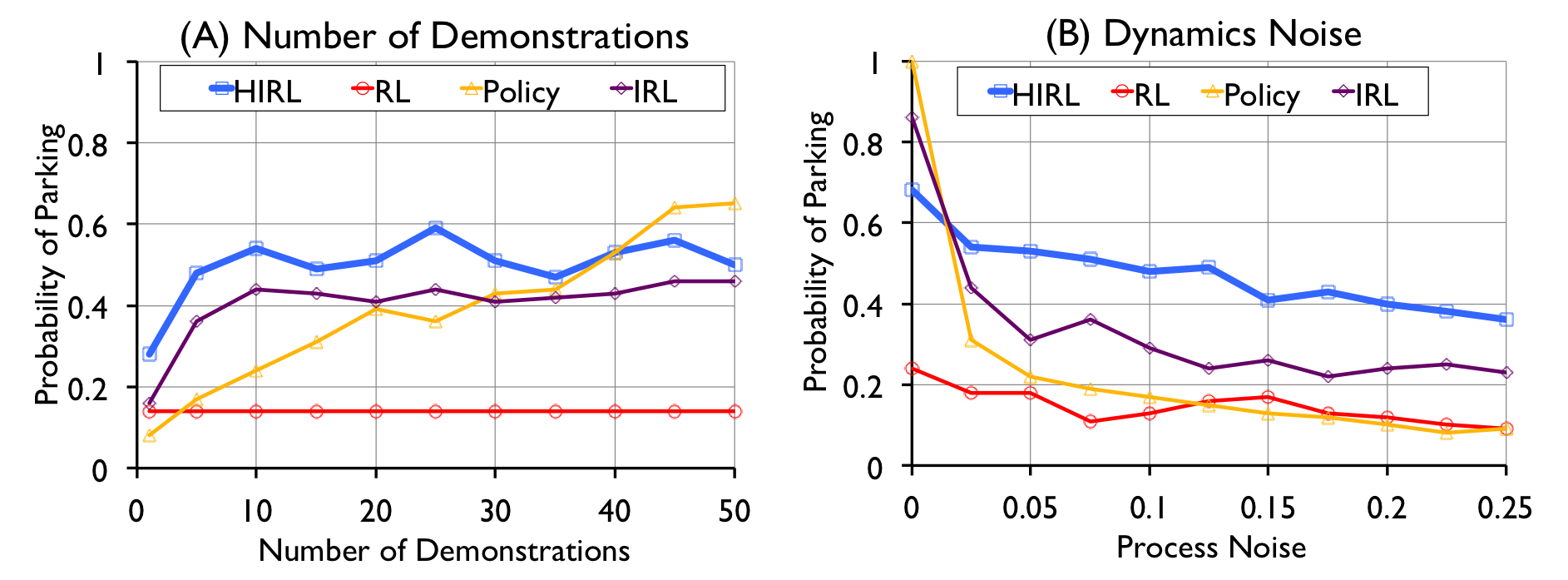}
 \caption{For a fixed number of RL episodes, we measure the success of the policies learned for each of the alternatives (Section \ref{exp:ppr}). [A] We vary the number of demonstrations provided to the different techniques and measure the success probability of the policies learned. \hirl has a higher success rate than the alternatives for a small number of demonstrations. [B] When there is no process noise, directly applying an open-loop policy works. However, we find that \hirl is more robust than the alternatives.  \label{exp:rcsegmentation-res2}}
\end{figure}

\begin{figure}[t]
\centering
 \includegraphics[width=\columnwidth]{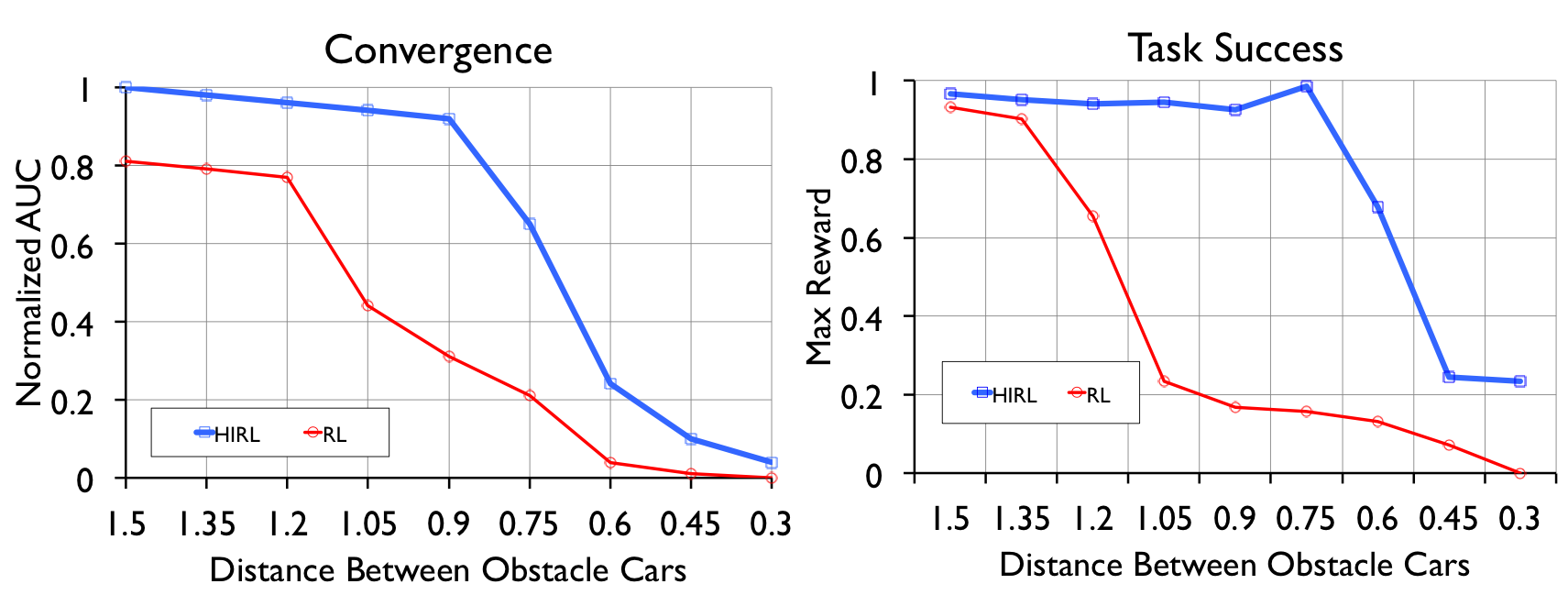}
 \caption{For a fixed number of RL episodes and 5 demonstrations, we plot the performance of \hirl as a function of the distance between the obstacle parked cars (Section \ref{exp:ppd}). 
 When the distance is large (the task is easier) the gap between \hirl and RL is smaller, however, as the task becomes harder the subgoals are more valuable. \label{exp:rcenv}}
\end{figure}

\begin{figure}[t]
\centering
 \includegraphics[width=\columnwidth]{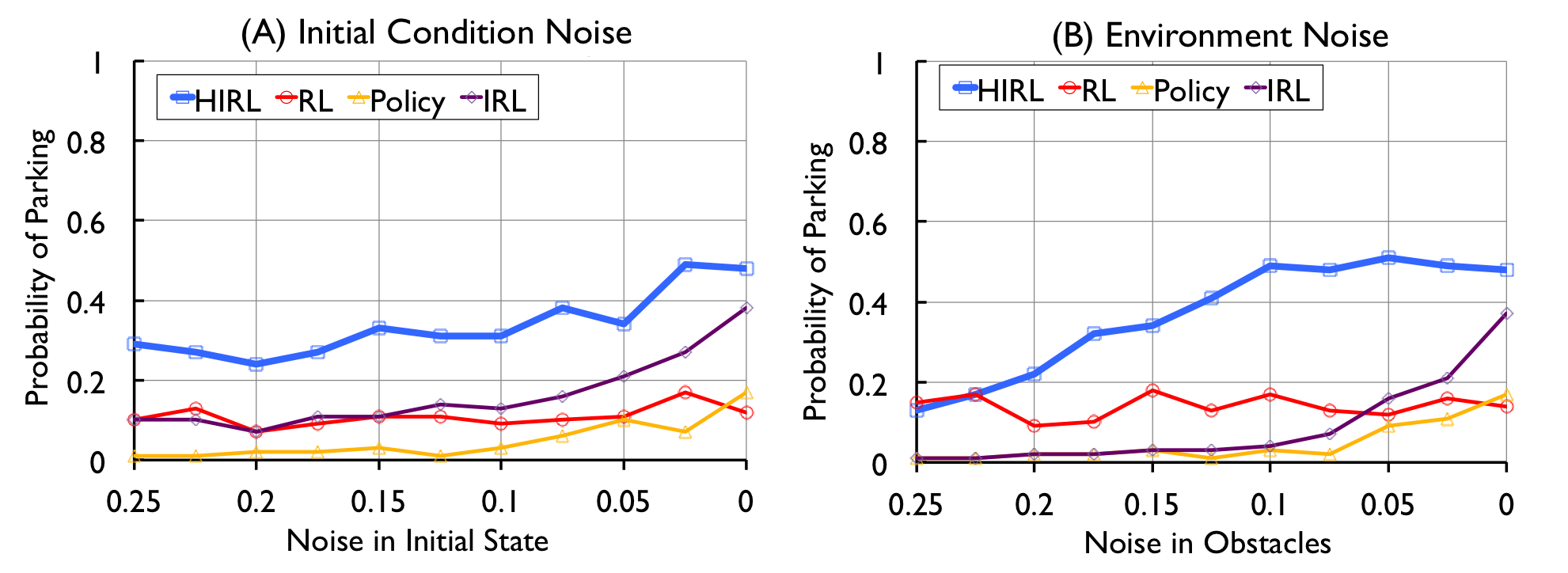}
 \caption{We collected 5 demonstrations in one environment and applied \hirl, the baseline, and policy learning, and measured the success rate of the learned policies for a fixed number of RL iterations. We evaluated these policies in a randomly perturbed  environment (Section \ref{exp:pper}). Noise is listed in terms of $\sigma$.\label{exp:rcenv2}}
\end{figure}

\subsection{Parallel Parking}\label{exp:pp}
We constructed a parallel parking scenario with an agent with non-holonomic dynamics and two obstacles. The agent can control its speed ($\|\dot{x}\|+\|\dot{y}\|$) and heading, and observe its x position, y position, orientation, and speed in a global coordinate frame.
If the agent parks between the obstacles, i.e., 0 velocity within a $15^\circ$ tolerance, the task is a success and the agent receives a reward of $1$. 
The agent's dynamics are noisy and with probability 0.1 will randomly add or subtract $5^\circ$ degrees from the steering angle.
If the agent collides with one of the obstacle or does not park in 200 timesteps the episode ends.
Next, we made the  Parallel Parking domain a little harder. We hid the velocity state from the agent, so the agent only sees $(x,y,\theta)$. As before, if the agent collides with one of the obstacle or does not park in 200 timesteps the episode ends.
We call this domain Parallel Parking-PO.

We collected 5 demonstrations and applied \hirl to learn the segments.
Figure \ref{exp:rcsegmentation} illustrates the demonstrations and the learned segments. There are two intermediate goals corresponding to positioning the car and orienting the car correctly before reversing.

\subsubsection{Convergence}\label{exp:ppc} In the first experiment, we use these learned segments to construct rewards in both the fully observed and partially observed problems (Figure \ref{exp:rcsegmentation-res}). 
In the fully observed problem, compared to IRL, \hirl converges to a policy with a 60\% success rate with about 3x less time-steps of exploration.
After 1250 episodes, the policy learned with \hirl has an 93\% success rate in comparison to a 60\% success rate for the baseline.
Also, for the same number of demonstrations, directly learning a policy gives a success rate of 17\%.

In the partial observation problem, there is no longer a stationary policy that can achieve the reward.
The learned segments help disambiguate dependence on history.
After 2000 episodes, the policy learned with \hirl has an 86\% success rate in comparison to a <10\% success rate for the baseline and the policy learning.

\subsubsection{Demonstrations and Robustness to Process Noise}\label{exp:ppr} Of course, the direct policy learning approach will work well if there is no stochasticity in the environment (i.e., open-loop control).
We evaluate these tradeoffs in the next experiment (Figure \ref{exp:rcsegmentation-res2}). 
If there is no process noise, the policy learning approach works perfectly for a fixed 5 demonstrations.
However even after adding a small amount of process noise, we find that its accuracy immediately drops.
Furthermore, we find that to achieve the same level of accuracy as \hirl after 250 episodes of exploration, need 40 demonstrations for the policy learning approach.

\subsubsection{Task Difficulty}\label{exp:ppd} Next, we explored the benefits of segmentation as a function of the hardness of the task.
We varied the distance between the obstacle cars and measured the performance of \hirl, where a smaller distance would make the task harder.
In comparison to the baseline, we found that segmentation is most beneficial when the task is harder (Figure \ref{exp:rcenv}). 

\begin{figure}[t]
\centering
 \includegraphics[width=\columnwidth]{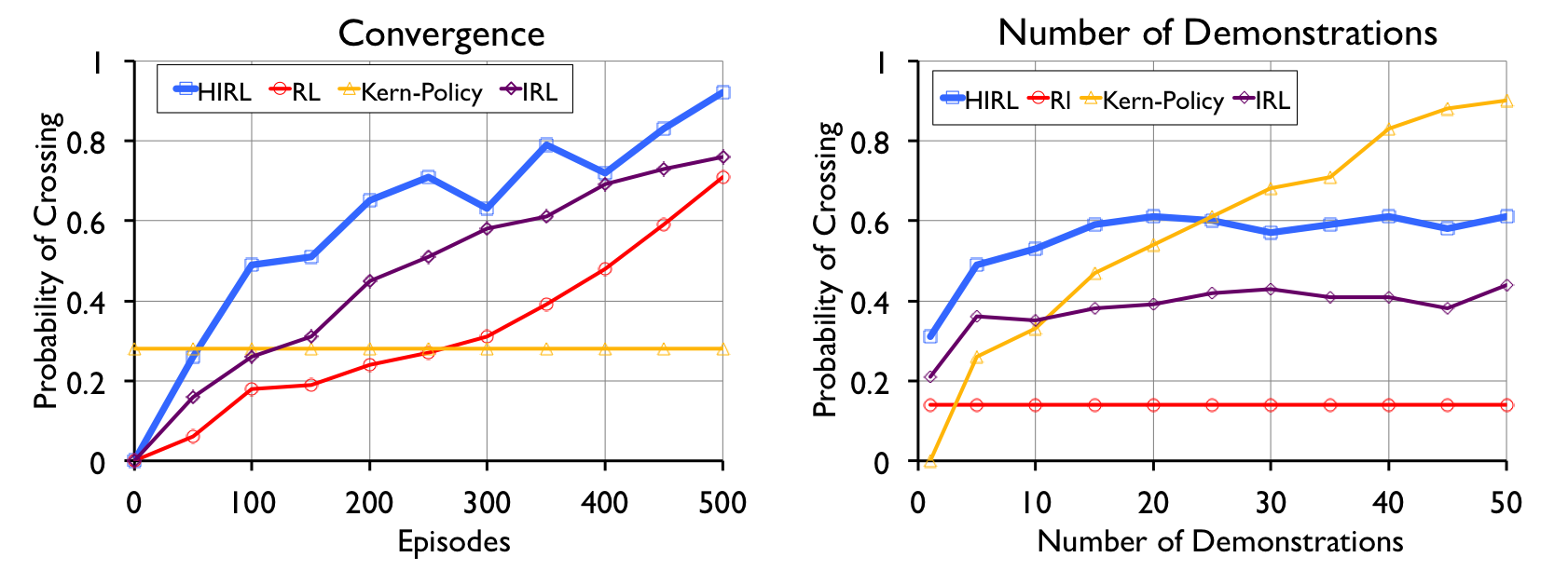}
 \caption{We measured the convergence of rewards constructed with \hirl and the alternatives (Section \ref{exp:acrobot}).
 [A] the success rate of the learned policy for 5 demonstrations, and varying the number of RL episodes. 
 [B] the success rate of the learned policy for a fixed number of episodes and varying the number of demonstrations.  \label{exp:acsegmentation-res2}}
\end{figure}

\subsubsection{Robustness to Environment Noise}\label{exp:pper} Finally, we explored the robustness of the learned policies to changes in the environment and starting position (Figure \ref{exp:rcenv2}).
We collected 5 demonstrations in one environment and applied \hirl, the baseline, and policy learning.
Then, we randomly perturbed the environment and evaluated the success rate of each policy.
We found that the policies learned with segmentation were very robust to noise in the initial state (i.e., the starting pose of the car).
On the other hand, the directly learned policies are not as robust to such noise.
While the actions at every state may change, the sub-goals stay the same.
We also found that the policies learned with segmentation were robust to noise in the obstacle car pose up to a point.

\subsection{Acrobot}\label{exp:acrobot}
This domain consists of a two-link pendulum with gravity and with torque controls on the joint. The dynamics are noisy and there are limits on the applied torque. The agent has 1000 timesteps to raise the arm above horizontal ($y=1$ in the images). If the task is a success and the agent receives a reward of $1$. 
Thus, the expected reward is equivalent to the probability that the current policy will successfully raise the arm above horizontal.
We generated $N=5$ demonstrations for the Acrobot task and applied segmentation. 
In Figure \ref{exp:acsegmentation-res2}, we plot the convergence of the all of the approaches.
We include a comparison between a Linear Multiclass SVM and a Kernelized Multiclass SVM for the policy learning alternative.
As before, we find that \hirl requires less demonstrations to converge to a more reliable policy.
\hirl converges 2.5x faster to a policy with a success rate of 60\%, and the direct policy learning outperforms \hirl for a fixed 100000 steps only after 25 demonstrations.

\begin{figure*}[t]
\centering
 \includegraphics[width=\textwidth]{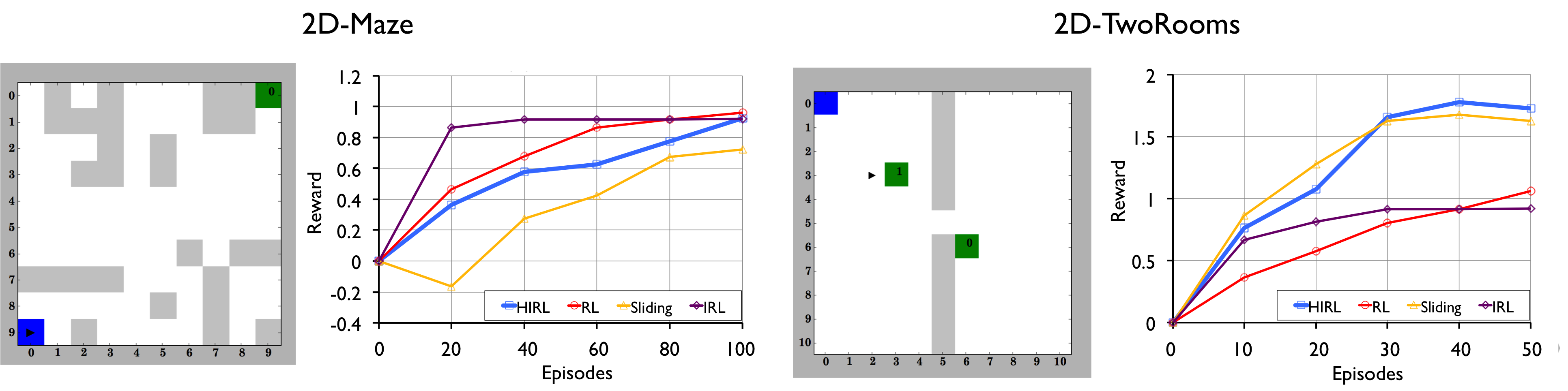}
 \caption{We summarize the results on two further domains (Section \ref{exp:further}). In the Maze domain, an agent plans a path to a single goal. We find that rewards constructed with IRL converge the fastest on this task. In the Two-Rooms domain, we find that alternative state-augmentation techniques ensure convergence, i.e., we can use a sliding window of states. It may not be possible to hard-code such augmentations in all cases, and \hirl provides a general framework to learn subtasks and augment the state-space with the appropriate variables.}
 \label{exp:gw2}
\end{figure*}

\subsection{Further Experiments}\label{exp:further}
Next, we highlight two additional scenarios that we constructed to evaluate \hirl.

\subsubsection{2D-Maze} In this domain, the agent solves a 2D Maze in a 11x10 grid (Figure \ref{exp:gw2}). There is no noise in the dynamics and there is a unique free path from start to end.
The MW domain was constructed to illustrate that over-segmenting a unique solution path can slow down convergence.
There is a single unique path that solves the maze, and this path is essentially revealed with IRL. Adding segments in addition to IRL does not improver performance, and in fact, it converges 60\% slower.
This domain was constructed as a counter-example to illustrate that \hirl is not always strictly better than the alternatives.

\subsubsection{2D-Two-Rooms} The two-rooms domains is another grid based environment (Figure \ref{exp:gw2}). 
Where there is a grid that is partitioned by a line of obstacles with a narrow opening. 
The agent has to collect a reward in the second room and return to the starting location.
The two goals are relatively close together.
The agent has 50 timesteps to achieve success.

We designed this domain in such a way that a sliding window of previous states could address the sequentiality problem.
Since the rooms are close together a sliding window states is sufficient to know whether the first goal was reached.
We find that \hirl can learn a reliable policy without hard-coding this sliding window, and achieves within 7\% convergence rate of the sliding window solution.

\subsection{Summary}
Table \ref{tab:my_label} summarizes the results of our experiments in terms of convergence rate and maximum attained reward on the Parallel Parking domain (with and without partial observation), Acrobot domain, and the 2-D motion planning domains.
In the appendix, we also provide details on the counter-examples that we constructed and their properties.

\begin{table*}[ht!]
\scriptsize
    \centering
    \begin{tabular}{|r|r|r|r|r|r|r|r|r|r|r|r|r|}
    \hline
         &  \multicolumn{2}{c|}{2D-MP-1}&
         \multicolumn{2}{c|}{2D-MP-2}& \multicolumn{2}{c|}{Two-Rooms} & 
         \multicolumn{2}{c|}{RC(FO)}
         & \multicolumn{2}{c|}{RC(PO)}& \multicolumn{2}{c|}{Acrobot} \\ 
    \hline
       & Max & AUC & Max & AUC & Max & AUC & Max & AUC & Max & AUC & Max & AUC\\
    \hline
        RL & $0.984$ & $10.976$ &$0.861$  & $15.440$& $1.090$ & $16.270$& $0.911$ &$109.76$ & $0.311$ &$27.419$ & $0.944$ &$3.447$\\
    \hline
        IRL & $0.987$ & $299.556$ & $0.861$ & $16.956$ & $0.759$ & $16.270$ & $0.950$ &$299.556$ & $0.444$ &$33.128$ & $0.920$ &$44.111$\\
    \hline
        TSC+Endpoints & $1.830$  & $322.125$ & $1.764$ & $14.07$ & $\mathbf{1.751}$& $\mathbf{18.953}$ & $\mathbf{0.991}$ &$164.127$ & $0.934$ &$123.115$ & $0.906$ &$20.935$\\
    \hline
        \hirl & $\mathbf{1.835}$ & $\mathbf{514.113}$ & $\mathbf{1.827}$ & $\mathbf{28.632}$ & $1.577$ & $17.141$ & $0.965$ & $\mathbf{514.113}$ & $\mathbf{0.958}$ & $\mathbf{333.897}$ & $\textbf{0.987}$ &$\textbf{65.512}$\\
    \hline
    \end{tabular}
    \caption{This table summarizes the convergence rate (\textbf{AUC}) and max reward (\textbf{MAX}) attained by a Q-learning agent using different reward and state-space representations. The RL agent directly optimizes the given reward for the task, IRL applies IRL without segmentation or state-space augmentation, TSC+Endpoints applies segmentation without IRL, and \hirl is our proposed technique. In all but one of the examples, \hirl converges faster than the alternatives. The exception is the Two-Rooms domain in which segmentation is still beneficial, but segmentation combined with IRL does not give a significant improvement over TSC+Segmentation.
    \label{tab:my_label}}
    
\end{table*}


\section{Future Work}
Our experimental results are very promising as they suggest that segmentation can indeed improve convergence in RL problems.
We believe that the proposed \hirl is a special case of a broader problem of learning hierarchies in MDPs.
One avenue for future work is modeling complex tasks as hierarchies of MDPs, namely, tasks composed of multiple MDPs that switch upon certain states and the switching dynamics can be modeled as another MDP. 

\subsection{Sub-Task Model}
We formalize the local problems studied in this paper in terms of MDPs.
We can generalize the definition of an MDP to allow for indefinite time-horizons.
Let $\mathcal{M}$ be an MDP as before, but instead of a finite-time horizon let $T:S\mapsto\{true,false\}$ be a stopping rule: 
\[\mathcal{M}=\langle S,A,P(\cdot,\cdot),R,T(\cdot)\rangle\]
Instead of executing a policy for a fixed T steps, the MDP executes until $T(\cdot)$ is true. 
The Linear-Gaussian dynamics in this paper are a special case of this idea.

A set of sub-tasks defines a universe $\mathcal{U} = \{\mathcal{M}_1,...,\mathcal{M}_k\}$ if they  are defined over the same state-space, action-space, and have the same dynamics. Within a universe, the only things that vary between sub-tasks are the reward functions and stopping rules
$\mathcal{M}_{i}=\langle R_{i},T_{i}\rangle$.

\subsection{Composite Task Model}
Given a universe of sub-tasks $\mathcal{U}$, a composite task is itself an MDP.
The current state of a composite task is a sub-task which is currently active and the termination state of the previous sub-task, so the state space $\mathbf{S} = \mathcal{U} \times S$.
The action space for the composite task is the set of policies $\mathcal{U}$ which is a next sub-task to attempt.
There is a transition function $P_{c}$ that given a current sub-task will transition to another sub-task in with some probability $\mathcal{U}$ conditioned on the outcome of the executed sub-task (the terminal state).
There is also a global reward $R_{c}$ and a global termination condition $T_{c}$:
\[
\mathcal{C} = \langle \mathbb{S}, \mathcal{U}, P_{c}, R_{c}, T_{c} \rangle
\]
This paper only studied sequential tasks with deterministic progressions, but  there are more opportunities to model interesting hierarchies with stochastic and non-sequential heirarchies.

\section{Conclusion}
Partitioning a task into sub-tasks with shorter-term rewards can lead to faster convergence to successful policies, but the challenge is defining the correct sub-task abstractions.
This paper explored a model for learning this partitioning from a set supervisor demonstrations.
We proposed framework \hirlfull (\hirl), and it addresses two problems: (1) given a set of featurized demonstration trajectories learn the locally linear sub-tasks, (2) use the learned subtasks to construct additional features for use in IRL.
We evaluate \hirl on 7 different domains with varying levels of non-linearity, stochasticity, partial observation, and state-space dimensionality.
We find that rewards constructed with \hirl converge the fastest in comparison to the alternatives (up-to $6$x faster than second best): IRL without segmentation, RL with a default delayed reward, and augmenting the state-space with a sliding window memory.
For the domains with known ground truth, we found that \hirl was within 10\% of the max reward achieved to \emph{a priori} perfect knowledge.

{\footnotesize 
\noindent \textbf{Acknowledgements:}
We would like to thank Anca Dragan, Stuart Russell, and Pieter Abbeel for discussions about this work.
We would also like to thank Daniel Seita, Jeff Mahler, and Michael Laskey for their feedback on early drafts.
This research was performed with UC Berkeley's Automation Sciences Lab under the UC Berkeley Center for Information Technology in the Interest of Society (CITRIS)``People and Robots'' Initiative http://robotics.citris-uc.org, and 
UC Berkeley's Algorithms, Machines, and People Lab.}



\bibliographystyle{abbrv}
\bibliography{deepP2P}

\appendix
\section{Appendix}

\subsection{Sequence of Stable Feedback Controllers}\label{sec:appendix1}
The proposed model naturally arises from a system controlled with linear state feedback controllers to the centroids of the $k$ target regions $[\rho_1,...,\rho_k]$. 
In the Transition State model,  $[\rho_1,...,\rho_k]$ are defined as the sublevel sets of multivariate Gaussian distributions.
For each of the Gaussian mixture components, let $[\mu_1,...,\mu_k]$ denote the respective expectations and $[\Sigma_1,...,\Sigma_k]$ denote the respective covariances.
We can show that the Transition State Clustering model naturally follows from of a sequence of stable linear full-state feedback controllers sequentially controlling the system to each $\mu_i$ (up-to some tolerance defined by $\alpha$).

Suppose, we model the agent's trajectory in feature space as a linear dynamical system with a fixed dynamics.
Let $A_r$ model the agent's linear dynamics and $B_r$ model the agent's control matrix:
\[
\mathbf{x}(t+1) = A_r\mathbf{x}(t) + B_r\mathbf{u}(t) + W(t).
\]
For a particular mixture component $i$, the agent applies a linear feedback controller with gain $C_i$, regulating around the target state $\mu_i$.
This can be represented as the following system (by setting $u(t)=-C_i\hat{\mathbf{x}}$):
\[
\hat{\mathbf{x}}(t) = \mathbf{x}(t) - \mu_i.
\]
\[
\hat{\mathbf{x}}(t+1) = (A_r-B_rC_i)\hat{\mathbf{x}}(t)+ W(t).
\]
If this system is stable, it will converge to the state $\hat{\mathbf{x}}(t) = \mathbf{0}$ which is  $\mathbf{x}(t) = \mu_i$ as $t \rightarrow \infty$.
However, since this is a finite time problem, we model a stopping condition, namely, the system is close enough to $\mathbf{0}$.
For some $z_\alpha$ (e.g., in 1 dimension 95\% quantiles are $Z_{5\%} = 1.96$):
\[
\hat{\mathbf{x}}(t)^T \Sigma^{-1}_i \hat{\mathbf{x}}(t) \le z_\alpha.
\]
If the agent's trajectory was modeled as a sequence $1...K$ of such controllers, we would observe the Transition State Clustering model with each $A_i = A_r-B_rC_i$, and the clusters would be an estimate of the $(\mu_i, \Sigma_i)$.

\subsection{Mixture Models And Linear Systems}\label{sec:appendix2}
Using a GMM to detect switches in local linearity is an approximate algorithm that has been applied in a number of prior works~\cite{moldovan2013dirichlet,calinon2014task, khansari2011learning}.
This is akin to a using a Gaussian kernel for kernelized change point detection~\cite{harchaoui2009kernel}.
We provide some intuition on why this model is sensible for our application.

Consider the following dynamical system:
\[
x_{t+1} = f(x_{t}) + w_{t}
\]
where $w_{t}$ is unit-variance i.i.d Gaussian noise $N(0,I)$.
Let us first focus on linear systems.
If $f$ is linear, then the problem of learning $f$ reduces to linear regression:
\[
\arg\min_{A} \sum_{t=1}^{T-1} \|Ax_{t} - x_{t+1}\|.
\]
Alternatively, we can think about this linear regression probabilistically.
Let us first consider the following proposition:

\vspace{0.75em}

\begin{proposition}
Consider the one-step dynamics of a linear system.
Let $x_{t} \sim N(\mu, \Sigma)$, then $\binom{x_{t}}{x_{t+1}}$ is a multivariate Gaussian.
\end{proposition}
\begin{proof}
This follows from the fact that $x_{t+1}$ can be expressed as a linear combination of independent multivariate Gaussian random variables.
\end{proof}

\vspace{0.5em}

Following from this idea, if we let $p$ define a distribution over $x_{t+1}$ and $x_{t}$:
\[
p(x_{t+1},x_{t}) \sim Normal
\]
For multivariate Gaussians the conditional expectation is a linear estimate, and we can see that it is equivalent to the regression above:
\[
\arg\min_{A} \sum_{t=1}^{T-1} \|Ax_{t} - x_{t+1}\| = \mathbf{E}[x_{t+1} \mid x_{t}].
\]

The GMM model allows us to extend this line of reasoning to consider more complicated $f$.
If $f$ is non-linear $p$ will almost certainly not be Gaussian.
However, GMM models can model complex distributions in terms of Gaussian Mixture Components:
\[
p(x_{t+1},x_{t}) \sim GMM(k)
\]
where $k$ denotes the number of mixture components.
The interesting part about this mixture distribution is that locally, it models the dynamics as before.
Conditioned on particular Gaussian component $i$ the conditional expectation is:
\[
\mathbf{E}[x_{t+1} \mid x_{t}, i \in 1...k].
\]
As before, conditional expectations of Gaussian random variables are linear, with some additional weighting $\phi(i \mid x_{t},x_{t+1})$:
\[
\arg\min_{A_i} \sum_{t=1}^{T-1} \phi(i \mid x_{t},x_{t+1}) \cdot \|A_ix_{t} - x_{t+1}\|.
\]
Every tuple $(x_{t+1},x_{t})$ probability $\phi(i \mid x_{t},x_{t+1})$ of belonging to each $i$th component, and this can be thought of as a likelihood of belonging to a given locally linear model.

\subsection{Alternative Approaches}
We consider the following alternative approaches to compare against \hirl. 

\subsubsection{RL}
This approach considers no segmentation and no history. It directly applies forward RL to the apparent state-space and uses a distance-to-goal reward function.

\subsubsection{Sliding Window}
This approach considers no segmentation but includes a sliding window of $k$ previous states in the state-space. It directly applies forward RL to the augmented state-space and uses a distance-to-goal reward function.

\subsubsection{IRL}
This approach uses MaxEnt-IRL to learn a reward function without segmentation and requires $N=5$ demonstrations. We apply forward RL to the learned reward function. 

\subsubsection{Endpoint Model}
This is a simplified approach to construct rewards using the learned $G$.
Let $\{\mu_1,...,\mu_k\}$ be the set of all of the means of $G$ learned with the algorithm in the previous section.
These means are in the feature space $\mathbb{R}^p$.
Let $\gamma$ denote the current progress of the task, i.e., the previously achieved goal  + 1. 
We can define a reward function as follows:
\[
R(s,a) = -\|f(s,a) - \mu_{\gamma}\|_2^2 
\]

\subsection{Counter-examples}
We constructed two scenarios, a Maze and a Pinball domain, in which \hirl actually performs worse than the alternatives. In the Maze domain, there is a 2D grid with a single unique solution path to a goal state.
In this problem, the segments found by \hirl provide no additional information compared to IRL.
In the Pinball domain, there is a ball on a table with obstacles that is moved by tilting the table.
The ball has elastic collisions with the obstacles and has noisy dynamics.
In this domain, we find that \hirl tends to over-segment this problem since every collision results in another linear regime. 

\begin{table}[ht]
\scriptsize
    \centering
    \begin{tabular}{|r|r|r|r|r|}
    \hline
         &\multicolumn{2}{c|}{Maze}& 
         \multicolumn{2}{c|}{Pinball}\\
    \hline
       & Max & AUC & Max & AUC\\
    \hline
        RL &  $\mathbf{0.960}$ & $2.575$& $0.481$ &$6.941$\\
    \hline
        IRL & $0.914$ & $\mathbf{3.575}$ & $0.424$ &$\mathbf{10.904}$\\
    \hline
        TSC+Endpoints & $0.944$ & $-0.448$&  $\mathbf{0.793}$ & $9.315$\\
    \hline
        \hirl & $0.924$ & $1.448$ & $0.722$ &$8.331$\\
    \hline
    \end{tabular}
    \caption{This table summarizes the convergence rate and max reward attained by a Q-learning agent using different reward and state-space representations on domains that were constructed to be counter-examples. \hirl does not perform as well in domains where there is a single path to the goal state. In this case, IRL finds the path and the additional states added by \hirl can actually impede convergence. }
    \label{tab:my_label}
\end{table}

\end{document}